\documentclass[letterpaper, 10 pt, journal, twoside]{IEEEtran}

\usepackage{graphicx} 
\usepackage{amsmath} 
\usepackage{amsfonts}
\usepackage{nicefrac}
\usepackage{microtype}
\usepackage{diagbox}
\usepackage{subfig}		
\usepackage{tikz}
\usepackage{adjustbox}
\usepackage{enumitem}
\usepackage{multirow}
\usepackage{booktabs}	
\usepackage{listings}
\usepackage{algorithm}
\usepackage{url}
\usepackage{flushend}
\usepackage[noend]{algorithmic}
\usepackage[noadjust]{cite}
\usepackage{xcolor}
\usepackage{comment}

\newcommand{\fillbox}[3]
{\bgroup
 \dimen1=#1\relax
 \dimen2=#2\relax
 \sbox0{\includegraphics[width=#1]{#3}}%
 \ifdim\ht0>\dimen2
 \dimen0=\dimexpr \ht0-\dimen2\relax
 \adjustbox{clip=true, trim=0pt 0.5\dimen0 0pt 0.5\dimen0}{\usebox0}%
 \else
 \sbox0{\includegraphics[height=#2]{#3}}%
 \ifdim\wd0>\dimen1
  \dimen0=\dimexpr \wd0-\dimen1\relax
  \adjustbox{clip=true, trim=0.5\dimen0 0pt 0.5\dimen0 0pt}{\usebox0}%
 \else
  \usebox0
 \fi
 \fi
\egroup}

\newtheorem{definition}{Definition}

\DeclareMathOperator*{\minimize}{minimize}

\usetikzlibrary{arrows, shapes, automata, petri, positioning}

\tikzset{
 place/.style={
  circle, 
  thick, 
  draw=blue!75, 
  fill=blue!20, 
  minimum size=6mm, 
 }, 
 transitionH/.style={
  rectangle, 
  thick, 
  fill=black, 
  minimum width=8mm, 
  inner ysep=2pt
 }, 
 transitionV/.style={
  rectangle, 
  thick, 
  fill=black, 
  minimum height=8mm, 
  inner xsep=2pt
 }
}

\definecolor{nero}{rgb}{0, 0, 0}
\definecolor{rosso}{rgb}{0.9, 0, 0}
\definecolor{verde}{rgb}{0, 0.6, 0}
\definecolor{blu}{rgb}{0, 0, 0.9}
\definecolor{grigio}{rgb}{0.52, 0.52, 0.51}

\newcommand{\CommonStyleCmd}[1]{{\small\textit{\textsc{#1}}}}

\newcommand{\Task}{\CommonStyleCmd{Task}}
\newcommand{\Tasks}{\CommonStyleCmd{Tasks}}

\newcommand{\RobotConfiguration}{\CommonStyleCmd{Robot\hspace{0.2em}Configuration}}
\newcommand{\RobotConfigurations}{\CommonStyleCmd{Robot\hspace{0.2em}Configurations}}

\newcommand{\RobotMovement}{\CommonStyleCmd{Robot\hspace{0.2em}Movement}}
\newcommand{\RobotMovements}{\CommonStyleCmd{Robot\hspace{0.2em}Movements}}

\newcommand{\Action}{\textsc{\CommonStyleCmd{Action}}}
\newcommand{\Actions}{\textsc{\CommonStyleCmd{Actions}}}

\newcommand{\vv}[1]{``#1''}
\newcommand{\ie}{{\em i.e.}}
\newcommand{\eg}{{\em e.g.}}

\newcommand{\FTL}{\mathit{FTL}}

\title{Optimal task and motion planning and execution for multi-agent systems in dynamic environments}

\author{
Marco Faroni$^{1}$, Alessandro Umbrico$^{2}$, Manuel Beschi$^{1, 3}$, 
Andrea Orlandini$^{2}$, Amedeo Cesta$^{2}$, Nicola Pedrocchi$^{1}$ 
\thanks{This work is partially supported by ShareWork project (H2020, European Commission: G.A. 820807). Authors from ISTC-CNR are also partially supported by ROXANNE project
(H2020, ROSIN – G.A. 732287).}

\thanks{$^{1}$ STIIMA-CNR: Institute of Intelligent Industrial Technologies and Systems, National Research Council of Italy 
 {\tt\small \{name.surname\}@stiima.cnr.it}}%
\thanks{$^{2}$ ISTC-CNR: Institute of Cognitive Sciences and Technologies, National Research Council of Italy 
 {\tt\small \{name.surname\}@istc.cnr.it}}%
\thanks{$^{3}$ Dipartimento di Ingegneria Meccanica e Industriale, University of Brescia 
 {\tt\small \{name.surname\}@unibs.it}}%
}

\begin{document}

\maketitle

\begin{abstract}
Combining symbolic and geometric reasoning in multi-agent systems is a challenging task that involves planning, scheduling, and synchronization problems.
Existing works overlooked the variability of task duration and geometric feasibility that is intrinsic to these systems because of the interaction between agents and the environment.
We propose a combined task and motion planning approach to optimize sequencing, assignment, and execution of tasks under temporal and spatial variability.
The framework relies on decoupling tasks and actions, where an action is one possible geometric realization of a symbolic task. 
At the task level, timeline-based planning  deals with temporal constraints, duration variability, and synergic assignment of tasks.
At the action level, online motion planning plans for the actual movements dealing with environmental changes.
We demonstrate the approach effectiveness in a collaborative manufacturing scenario, in which a robotic arm and a human worker shall assemble a mosaic in the shortest time possible. 
Compared with existing works, our approach applies to a broader range of applications and reduces the execution time of the process.
\end{abstract}


\section{Introduction}
\label{sec: introduction}

\IEEEPARstart{H}{uman-robot} collaboration (HRC) boosts the flexibility of manufacturing processes, although the inefficient coordination between humans and robots often jeopardizes productivity \cite{T-CYB-survey-pHRC}.
From a planning perspective, efficiency in HRC is tied to different intertwined problems. 
First, the system should find a suitable sequence of operations (task planning), assign them to the agents (task assignment), and schedule their execution (scheduling). 
At run-time, the execution of the operations should be adapted to the human and robot's state (motion planning and replanning). 
All these steps should also consider the variability of the duration and good (or bad) synergy of simultaneous collaborative operations.
The complexity of the overall planning problem limits the effectiveness of existing methods in real-world scenarios, and 
standardized shared approaches to task and motion planning are still to come.

In this paper, we propose a tiered approach interleaving task planning, scheduling, assignment, and action planning for multi-agent systems.
The method addresses the tasks' temporal and geometric uncertainty by decoupling the abstract representation of the task from all its possible realizations.
Timeline-based planning reasons on abstract tasks, while online action planning optimizes their geometric implementations.
Compared with existing methods, our approach deals with a broader variety of real-world problems and reduces execution and idle times.

\subsection{Related works}
\label{sec: related-works}
This paper deals with interleaving task planning and motion planning and how to consider human behaviors in this process.
Existing methods address the first aspect by following the combined task and motion planning (TAMP) paradigm \cite{Lozano-Perez-TAMP-review}. 
Usually, TAMP provides a task planner able to reason geometrically through calls to a motion planning algorithm. In such a hierarchical approach, a task planning algorithm finds a feasible sequence of actions, and a motion planner checks for geometric feasibility \cite{srivastava2014combined, Lozano-Perez2014, Dantam-incremental-task-motion, Lozano-Perez-IJRR2018, SkiROS}.
Most TAMP methods focus on the feasibility of the plan rather than its optimality (except for a few exceptions~\cite{Toussaint-logic-geometric, Zhang2016}).
Few works address temporal planning to consider task duration \cite{Magazzeni-temporal-reasoning}.

HRC-oriented works usually focus on sub-problems such as scheduling human and robot actions \cite{Zanchettin-scheduling, HAM-Boeing777,Minto:task-allocation-ssm}, or cooperative planning at a symbolic level \cite{Tomizuka-human-aware-task-planning, wang-online-action-planning, Li-sequence-planning}.
Few works address TAMP by explicitly modeling the human agent \cite{T-CYB-human-agent-collaboration, T-CYB-human-agent-collaboration-part2}.
For example, \cite{Alami-HATP} and \cite{Alami-Lallement-HATP} proposed a hierarchical agent-based task planner, where complex tasks are decomposed into simple actions.
The method improves the collaborative experience by considering human preferences as social costs, but throughput-oriented objectives are not considered~\cite{Alami-using-human-knowledge, Alami-dealing-with-online-human}.
In manufacturing-oriented methods, \cite{Makris2020} optimizes the ergonomics of the human worker by using an online workflow scheduler. 
\cite{Casalino-interleaved-tamp} and \cite{Casalino-T-RO} proposed a TAMP framework for planning and executing tasks using first-order logic graphs.
A contingent-based approach was proposed in \cite{Rosell-knowledge-oriented-tamp}, and \cite{Rosell-Applied-Sciences} to deal with uncertainty on the outcome of actions.

The approaches above focus on finding feasible plans and do not consider: (i) process throughput; (ii) temporal constraints and uncertainty; (iii) human-robot synergy.
Regarding (i) and (ii), timeline-based task planning~\cite{hsts-94} has proved to be a powerful approach in many real-world applications \cite{goac-11, compint10mrspock}. 
The value of this approach consists of integrating planning and scheduling in a unified reasoning framework, making decisions about what actions to perform and when.
This approach can also model the \emph{features' controllability}, \ie, the planner knows whether it can control the task's beginning or the end~\cite{morris01}. 
Timelines were applied to HRC in~\cite{Pellegrinelli2017}, although the integration with motion planning was inefficient because it reasoned at a low level of abstraction (point-to-point movements). Consequently, motion plans were pre-computed, hindering the flexibility of the approach in dynamic environments.

\subsection{Contribution}\label{sec: contribution}
\begin{figure}
 \centering
 \includegraphics[height=8.5cm,width=0.85\columnwidth]{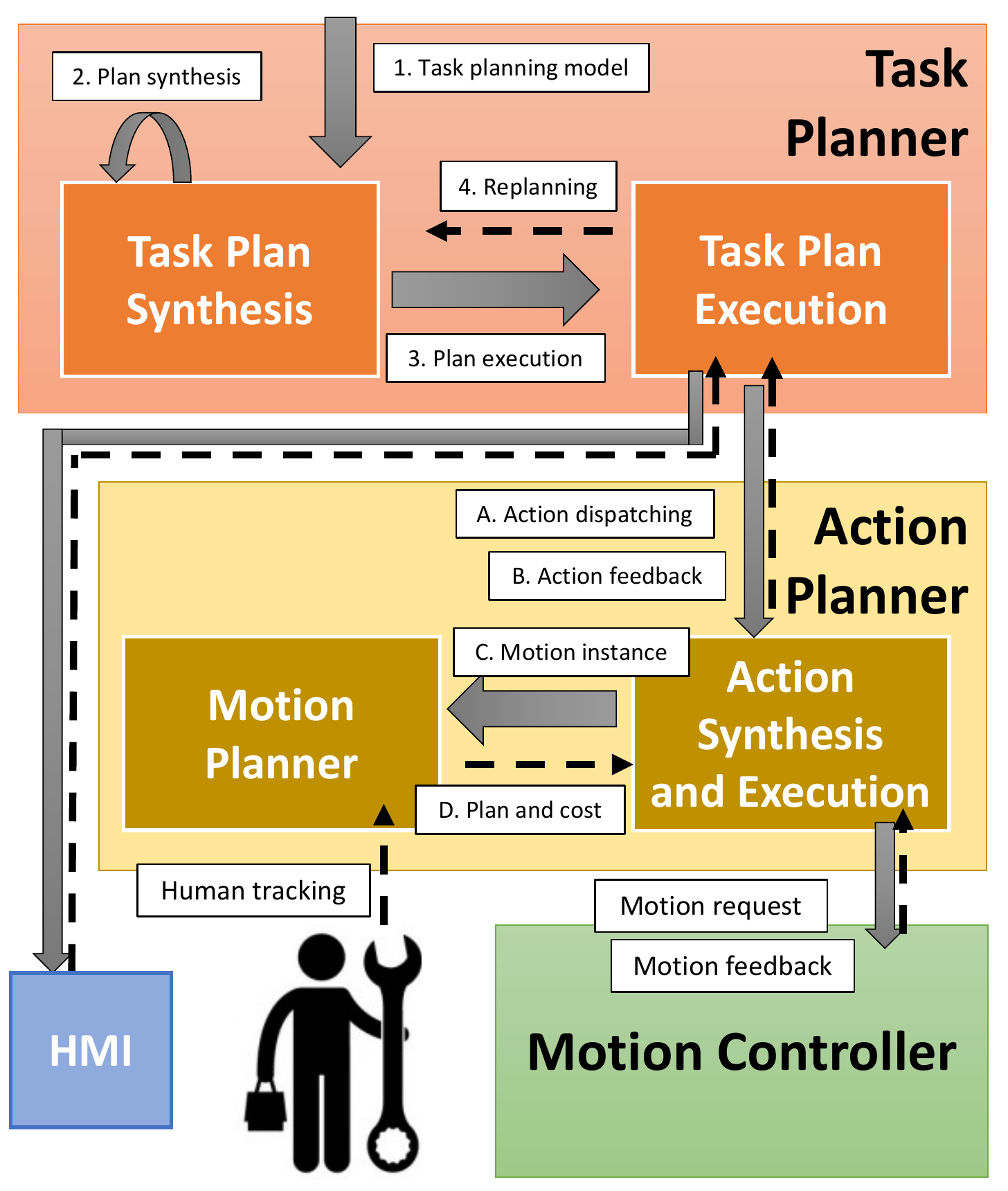}
 \caption{Proposed hierarchical framework.}
 \label{fig: integration}
\end{figure}

As shown in Figure~\ref{fig: integration}, we propose a hierarchical planner where the higher layer reasons over symbolic tasks, optimizing their order, scheduling, and assignment under temporal constraints and duration variability. At the same time, the lower level turns the symbols into robot motions, selecting the optimal task execution among all the possible alternatives.

Our formulation relies on the definition of \Actions\ as a set of instances of a \Task\ (Section~\ref{sec: approach}). Compared to previous works, such formulation explicitly maps a symbolic \Task\ to all its possible geometric realizations. 
From the task planning point of view, this formulation considers the duration uncertainty of the tasks, temporal constraints, and the possible synergy of simultaneous tasks performed by the different agents. 
At the action level, the robot plans and executes the its motions based on the current context.

Then, we propose algorithms to solve the proposed task and action planning problems (Section~\ref{sec: solvers}).
On the one hand, we convert the optimal task planning problem into a multi-objective problem -- which constitutes a novelty in timeline-based planning -- and use the notion of Pareto optimality to optimize the coupling of simultaneous tasks.
On the other hand, we make the optimal action planning problem tractable online by converting it into a multi-goal motion planning problem that can be solved with efficient off-the-shelf algorithms.

Compared to previous works, our formulation is robust to temporal and spatial uncertainty typical of hybrid multi-agent systems (\emph{e.g.}, human-robot collaboration).
We assess the broad applicability of the approach and its superiority to existing works qualitatively and experimentally (Sections~\ref{sec: qualitative-assessment} and~\ref{sec: case_study}), showing that our approach applies to a broader range of applications and reduces execution and idle times of the process.
A video of the experiments is attached to the manuscript.

\section{Task and action planning formalization}
\label{sec: approach}

\subsection{Definitions and Approach at a Glance}\label{sec: statement}
\noindent Our approach builds on the following definitions:
\begin{definition}[Worker]\label{def-worker}
It is an agent that performs a job assigned by the system. It can be a human worker or a robot.
\end{definition}
\begin{definition}[Task]\label{def-task}
It is a step of the production process performed by a worker. 
\Tasks\ do not provide geometrical information on their execution, as they model an operation at a higher level of abstraction. 
Each \Task\ might be performed in several ways, making its duration a random variable rather than a scalar variable.
\end{definition}

\begin{definition}[Robot Configuration]\label{def-configuration}
It is a vector collecting the joint positions and the auxiliaries' state\footnote{Example: consider a 2-axes mechanism with a gripper. The \textsc{Robot Configuration} is a vector of length equal to three, where the first two components are the joint positions, and the third one is the gripper state.}.
\end{definition}

\begin{definition}[Robot Movement]
it is a discrete change of the \RobotConfiguration. Each \RobotMovement\ might be realized by an infinite number of possible trajectories that satisfy the physical constraints of the robot.
\end{definition}

\begin{definition}[Action]\label{def-action} 
An \Action\ is a sequence of \RobotMovements. It is a feasible implementation of a \Task.
Each \RobotMovement\ is defined by the \Action\ parameters\footnote{Example: \textsc{Task} \vv{screw bolt A} corresponds to different \textsc{Actions}, each composed by the same sequence of \textsc{RobotMovements}: \vv{open the gripper}, \vv{move to grasp point $X$} (where $X$ is the position of a particular screwdriver), \vv{close the gripper}, \vv{move to $Y$} (where $Y$ is the screwing position for bolt A), \vv{screw bolt $A$}. Each \textsc{Action} differs in the value of $X$, $Y$, $A$. For example, multiple suitable screwdriver might be available in different locations.}.
\end{definition}


The proposed hierarchical approach is shown in Figure~\ref{fig: integration}. 
The task planner uses a planning model of the process to search for a feasible, possibly optimal sequence of \Tasks.
The planning model specifies the \Tasks\ to be performed with temporal and allocation constraints. 
The Task Plan Execution module dispatches the target \Task\ to the workers. The Action Planner converts each \Task\ into a set of feasible \Actions. 
Then, it chooses the best \Action\ among the feasible ones and sends the motion plan to the robot controller.
If the Action Planner fails at planning or executing (\emph{e.g.}, no trajectory could be computed or task execution fails at run-time), the Task Planner re-plans according to the Sense-Plan-Act paradigm~\cite{gat98}.

\subsection{Model Formalization}\label{sec: model formalization}
The following elements describe a collaborative process: 
\begin{description}
\item[$\mathcal{W} \,=\, \{H, R\}$] is the set of workers, \ie, a human and a robot; 
\item[$\mathcal{P}=\, \{p_i\}$] is the set of production targets;
\item[$\mathcal{T}=\, \{t^{p_i}_j\}$] is the set of \Tasks\ necessary to carry out a production target $p_i$;
\item[$\mathcal{A}=\, \{a_j\}$] is the set of \Actions;
\item[$D: \mathcal{T} \rightarrow \mathbb{R}^2$] is the duration function that associates a \Task\ with an interval $\left[d_{\mathrm{min}}, d_{\mathrm{max}}\right]$; 
\item[$T: \mathcal{T} \rightarrow \mathcal{T}$] is the transition function that defines valid transitions among \Tasks; 
\item[$\gamma: \mathcal{T} \rightarrow \{c, pc, uc\}$] is the controllability tag. The tag is ``controllable'', ``partially controllable'' or ``uncontrollable'' if the system can decide the execution start and end, only the start or neither of the \Task\ a worker performs.
\item[$F: \mathcal{T} \rightarrow \{\{H\}, \{R\}, \{H, R\}\}$] is a function that defines for each \Task\ which worker can execute it; 
\item[$S: \mathcal{A}\rightarrow \mathcal{T}$] is a function that maps each \Action\ to its corresponding \Task.
\item[$SV=\left( V, T, D, \gamma\right)$] is a ``state variable'' that describes the behaviors of a domain feature that is represented by the set of tasks $V=\left\{v_i\right\}\in \mathcal{T}$. Such a set gathers all the valid \Tasks\ that can be executed over time for that specific feature. $T$, $D$, $\gamma$ are defined as above. 
\item[$x=$]$\left(v, \left[e, e'\right], \left[d, d' \right], \gamma\left(v\right)\right)$ is a ``token'' where $v \in V$, $\left[e, e'\right]$ is  end-time interval, and  $\left[d, d' \right]=D(v)$ is the duration interval for task $v$.
\item[$\FTL_{SV}=\left\{x_j\right\}$] is a flexible timeline of a state variable $SV$ representing a temporal sequence of tokens $x_j$.
\end{description}
Solving a collaborative TAMP problem consists of identifying a task plan, a temporal schedule, and an assignment of the tasks considering that each task can be realized by a set of actions and each movement composing an action can be executed by an infinite set of trajectories. Each task plan is modeled through flexible timelines of tokens of state variables. 

\subsection{Task planning model for human-robot cooperation}
\label{subsec: problem-ps}

Given the notation in Section \ref{sec: model formalization}, we define the production goals, each worker's possible behaviors, and the synchronization rules to model a human-robot scenario.
Refer to \cite{marvel2015, tan2008hta} for a description of the formalization approach.

First, consider a set of high-level production targets $V^p= \{p_i\}$, where $p_i \in\mathcal{P}$. 
Each $p_i$ can be further associated with a set $V^{p_i} = \{t^{p_i}_{j}\}$ that gathers \Tasks\ to carry out $p_{i}$, where $t^{p_i}_j \in\mathcal{T}$ \footnote{The set $V^p$ could be 
$\{$\textit{assembly A, check the quality of B, disassemble C}$\}$, and $V^{p_1}=\{$\textit{Take the bottom of part A, pick and places the screws of A, Tight the screws of A}$\}$.}.
Second, we consider a generic worker $w_k \in\mathcal{W}$ that may implement some of the needed tasks in $V^{p_i}$, and denote by $V^{w_k}=\{t^{p_i, w_k}_m\}\subset V^{p_i}$ the subset of tasks $t^{p_i}_j$ that a worker $w_k$ can do according to $F$. 

Additionally to the set of tasks, we need to define their precedence constraints, task duration, and controllability. 

To gather all this information complactly, we denote by $SV^p = \bigl(V^p, T^P, D^p, \gamma^p \bigr)$ the production state variable associated with the high-level production targets $V^p = \{p_i\}$ and by $SV^{p_i} = \bigl(V^{p_i}, T^{p_i}, D^{p_i}, \gamma^{p_i} \bigr)$ the production state variables associated with the production tasks $V^{p_i} = \{t^{p_i}_j\}$ necessary for a particular production target $p_i$. 
The transitions $T^p$ and $T^{p_i}$ are usually an input of the model, while duration and controllability come from the worker modeling and selection (see below).
Then, we denote by $SV^{w_k} = \left( V^{w_k}, T^{w_k}, D^{w_k}, \gamma^{w_k} \right)$ as the behavior state variables for the $k$th worker (often one robot and one human). 
Transitions $T^{w_k}$, duration $D^{w_k}$, and controllability $\gamma^{w_k}$ are usually input of the model. 
The duration is an interval $D^{w_k}(t_m^{p_i, w_k}) = \bigl[\, d^{w_k}_m - \delta^{w_k}_m, \, d^{w_k}_m \, \, + \delta^{{w_k}}_m\, \bigr]$ that describes the duration uncertainty when worker $w_k$ performs $t^{p_i, w_k}_m \in\ V^{w_k}$. 
Controllability $\gamma^{w_k}\bigl(t^{p_i, w_k}_m\bigr)$ 
depends on the nature of the worker.
Task $t_m^{p_i, w_k}$ is uncontrollable if performed by humans since they may refuse to do a task or quit halfway through. 
Conversely, it is partially controllable if performed by a robots, as they may not be able to finish a task because humans obstruct all possible way-outs. 

As mentioned above, much information of the complete model is intertwined. 
For example, the execution time of the process task $t_j^{p_i}$ results 
\begin{align*}
D^{p_i}\bigl(t_j^{p_i}\bigr) = \left[\, \min_{t_m^{p_i, w_k}}\left(d^{w_k}_j - \delta^{w_k}_m\right), 
    \, \max_{t_m^{p_i, w_k}}\left(d^{w_k}_m \, \, + \delta^{{w_k}}_m\right)\, \right]\\
    \quad\mbox{s.t.}\quad{t_m^{p_i, w_k}}\,\mbox{implements}\;t_{j}^{p_i}
\end{align*}
otherwise, the problem is unfeasible. 
Similarly, the map of processes tasks $t_j^{p_i}$ into several feasible workers tasks of $t^{p_i, w_k}_i$ according to the different constraints (precedence constraints, resource constraints, \textit{etcetera}) is far to be simple. 
Proper SW tools autonomously compute such information (see Sec. \ref{sec: case_study}) that complete the model from user input, using proper synchronization rules $R$.

Given a set of state variables $SV^P$, $SV^{p_i}$, $SV^{w_k}$ and synchronization rules $R$, we can now introduce the timeline-based planning formalization \cite{hsts-94, cialdea2016planning}. 
We consider the generic state variable $SV^g= \bigl(V^g, T^g, D^g, \gamma^g \bigr)$ and denote by $x^g_j=\bigl(v^g_j, \bigl[{e^g_j}^-, {e^g_j}^+\bigr], \bigl[{d^g_j}^-, {{d^g}_j}^+\bigr], \gamma\bigl(v^g_j\bigr)\bigr)$ a ``token'' with $v^g_j \in V^g$, and $D^g\bigl(v^g_j\bigr)=\bigl[{d^g_j}^-, {{d^g}_j}^+\bigr]$. 
We define a flexible timeline  $\FTL_{SV^i}$ of a state variable $SV^i$ as a sequence of tokens $\{x_j^g\}$ that span the process execution time and describes what the worker does over the process.
Finally, a timeline-based plan $\pi$ consists of a set of flexible timelines, one for each state variable, valid with respect to $R$. A timeline-based solver exploits different search techniques to find the optimal $\pi$ among the feasible ones.

\begin{figure}[t]
  \centering
  \begin{tikzpicture} [
   auto, 
   decision/.style = { diamond, draw=blue, thick, 
        text width=5em, text badly centered, 
        inner sep=1pt, rounded corners }, 
   block/.style = { rectangle, draw=black, thick, 
         text width=3.cm, text centered, 
        rounded corners, minimum height=1em}, 
   line/.style  = { draw, thick, ->, shorten >=2pt }, 
   line_2/.style  = { draw, thick }, 
   ]
   
   \matrix [column sep=4mm, row sep=2mm] {
       & \node [text centered] (begin) {\footnotesize \textbf{begin}}; & \coordinate (point3){};& \coordinate (point4){};\\
       & \node [block] (first) {\footnotesize move to pre-picking pose}; \quad&
       & \node [block] (fifth) {\footnotesize move to post-grasp pose};\\
       & \node [block] (second) {\footnotesize open gripper}; \quad&
       & \node [block] (sixth) {\footnotesize move to pre-placing pose}; \\
       & \node [block] (third) {\footnotesize move to picking pose}; \quad&
       & \node [block] (seventh) {\footnotesize move to placing pose}; \\
       & \node [block] (fourth) {\footnotesize close gripper}; \quad&
       & \node [block] (eighth) {\footnotesize open gripper}; \\
       & \coordinate (point1){}; & \coordinate (point2){};&\node [text centered] (end) {\footnotesize \textbf{end}}; \\
   };
   
   \begin{scope} [every path/.style=line]
   \path (begin)  -: node {\footnotesize}(first);
   \path (first)  -: node {\footnotesize}(second);
   \path (second)  -: node {\footnotesize}(third);
   \path (third)  -: node {\footnotesize}(fourth);
   \path (point4)  -: node {\footnotesize}(fifth); 
   \path (fifth)  -: node {\footnotesize}(sixth);
   \path (sixth)  -: node {\footnotesize}(seventh);
   \path (seventh)  -: node {\footnotesize}(eighth);
   \path (eighth)  -: (end);
   \end{scope}
   \begin{scope} [every path/.style=line_2]
   \path (fourth)  -: node {\footnotesize}(point1);
   \path (point1)  -: node {\footnotesize}(point2); 
   \path (point2)  -: node {\footnotesize}(point3); 
   \path (point3)  -: node {\footnotesize}(point4); 
   \end{scope}
  \end{tikzpicture}
  \caption{Example: a pick-and-place action.}
  \label{fig: action_model}
 \end{figure}
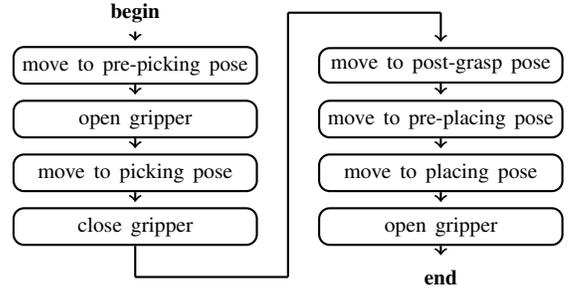
\subsection{Action planning model for human-robot cooperation}
\label{subsec: action-planning}
The action planner finds the best \Action\ to execute a given \Task\ $t_j^{w_k}$.
To do so, it gathers all the necessary geometric information from the scene descriptor (\eg, through queries to a database).
Then, it determines all the actions that can realize task $t_j^{w_k}$ (\ie, all actions $a_i \in \mathcal{A}$ such that $t_j^{w_k}=S(a_i)$, according to Definition \ref{def-action}).
For example, consider $t_j^{w_k}$ equal to ``pick a blue object and place it in an empty box''. 
The action planner uses the tags ``blue object'' and ``empty box'' to identify all the locations of the scene to which these labels are assigned.
Then, the action planner finds the best sequence of \RobotMovements\ that connects those locations in the order specified in Figure \ref{fig: action_model}.

We formulate the action planning problem as the identification of the best path on a directed graph $\mathcal{G}=(U_g, E_g)$, where $U_g$ is the set of vertices and $E_g$ is the set of edges, and:
\begin{itemize}
 \item $U_{g, i}$ is a set of \RobotConfigurations, that corresponds to the goal of the $i$th \RobotMovement;
 \item $E_{g, j}$ connects two sets of \RobotConfigurations, \ie, each edge is a set of \RobotMovements.
\end{itemize}
Indeed, $U_{g, i}$ includes all the \RobotConfigurations\ that are logically equivalent for the task (\eg, each one is the location of equivalent ``blue objects''). Furthermore, if the representation of the location is in the Cartesian space, many joint configurations may correspond to each Cartesian pose. 

For a generic action composed of $n$ movements, the set of vertices is therefore given by
$$U_g = \bigcup_{i=0}^n U_{g, i}$$
where the starting vertex $U_{g, 0}$ is a single \RobotConfiguration\ since the starting configuration is usually known.

The action planner shall find the shortest path on $\mathcal{G}$ from a vertex in $U_{g, 0}$ to a vertex in $U_{g, n}$.
The advantage of this formulation is that it maps a symbolic task into all its possible realizations.
Recalling the example of the blue cubes, the best action is optimal for all possible inverse kinematic solutions of all grasping points of all available blue cubes.

\section{Optimization and solvers}
\label{sec: solvers}

This section proposes algorithms to solve task and action planning problems despite the high computational complexity of these problems in real-world scenarios. Without loss of generality, we will refer to the case where the workers are one robot and one human.

\subsection{Optimization of collaborative processes}
\label{subsec: optimization-ps}

Consider a set of interaction windows $\mathcal{Y} = \{y_j\}$ during which the human and the robot perform some task $t_i^{p_k}$ associated with a production target $p_k$.
Let $b^{H}_{i, j}$ and $b^{R}_{i, j}$ be binary control variables such that:

\begin{equation}
\begin{aligned}
 b^{R}_{i, j} &= \left \{
 \begin{aligned}
 &1, && \text{if the robot does $t_i^{p_k}$ during $y_j$} \\
 &0, && \text{otherwise}
 \end{aligned} \right.\\
 b^{H}_{i, j} &= \left \{
 \begin{aligned}
 &1, && \text{if the human does $t_i^{p_k}$ during $y_j$}\\
 &0, && \text{otherwise.}
 \end{aligned}\right.
\end{aligned}
\end{equation} 
The robot and the operator
can perform only one task during a particular interaction window, namely:
\begin{equation}
\label{eq-logic.constraint1}
	\forall\ j\;\; \mbox{s.t.}\;\; y_j \in \mathcal{Y}, \quad \sum_{i} b^{R}_{i, j} = 1, \;\;\mbox{and}\;\;
	\sum_{i} b^{H}_{i, j} = 1
\end{equation}
Each task can be assigned only once during 
a process and thus executed only by the human or the robot, \ie,
\begin{equation}
\label{eq-logic.constraint3}
	\forall\ i\;\; \mbox{s.t.}\;\;t_i^{p_k} \in \mathcal{T}, \quad\sum_{j} b^{R}_{i, j} + b^{H}_{i, j} = 1
\end{equation}
A duration cost function $f_d$ can be therefore defined as:
\begin{equation}
\label{eq-duration-function}
f_d = \sum_{i} \sum_{j} d^R_i b^{R}_{i, j} + d^H_i b^{H}_{i, j}
\end{equation}
where $d^H_i$ and $d^R_i$ are the expected duration of 
task $t_i^{p_k} \in \mathcal{T}$ when performed by the human and by the robot separately.

However, \eqref{eq-duration-function} does not consider coupling effects between the robot and the human.
For example, if the robot and the human move to the same area concurrently, the robot will either stop or slow down for safety reasons.
To capture this synergy, we define $\Delta d^R_{i, j}$ and $\Delta d^H_{i, j}$ as
\begin{gather}
\label{eq-duration-coupled}
\Delta d^R_{i, j} = d^R_{i, j} - d^R_i~~~\text{and}~~~\Delta d^H_{i, j} = d^H_{i, j} - d^H_i
\end{gather}
where $d^R_{i, j}$ is the expected duration of task $t_i^{p_k}$ performed by the robot while the human is performing $t_j^{p_k}$ (and vice versa for $d^H_{i, j}$). 
The synergy cost function $f_s$ is therefore defined as:
\begin{align}
f_s = 
\sum_{i} \sum_{j} \sum_{k} \Delta d^R_{i, j} b^H_{j, k} b^R_{i, k} + \Delta d^H_{i, j} b^R_{j, k} b^H_{i, k} \nonumber\\ 
=
\sum_{i} \sum_{j} \left(s_{i, j} \sum_{k} b^H_{j, k} b^R_{i, k}\right)
\label{eq-synergy-function}
\end{align}
where $s_{i, j} = \Delta d^R_{i, j} + \Delta d^H_{i, j}$ is a synergy coefficient, \ie, an index of the simultaneous tasks coupling, as shown in Table \ref{table-synergy-matrix}.

In conclusion, the solution plan $\pi$ is a solution to the following multi-objective optimization problem:
\begin{equation}
\label{eq-multi-obj}
	\minimize_{\pi}\, \{ f_d, f_s \}
\end{equation}
subject to constraints \eqref{eq-logic.constraint1} and \eqref{eq-logic.constraint3}.
This paper focuses on time-efficiency criteria, but the extension to other objectives is straightforward and does not undermine the search strategy.

\begin{table}[t]
 \caption{Synergy matrix. Each element $s_{i, j}$ represents the increment or decrement of duration given by the simultaneous execution of tasks $i$ and $j$.}
 \centering
 \begin{tabular}{c | c c c c c c}
 \backslashbox{R}{H} & 0 & 1 & 2 & ... & n-1 & n\\\hline
 0 & $\infty$ & $s_{0, 1}$ & $s_{0, 2}$ & ... & $s_{0, n-1}$ & $s_{0, n}$\\
 1 & $s_{1, 0}$ & $\infty$ & $s_{1, 2}$ & ... & $s_{1, n-1}$ & $s_{1, n}$\\
 ... & ... & ... & ... & ... & ... & ...\\
 n-1 & $s_{n-1, 0}$ & $s_{n-1, 1}$ & $s_ {n-1, 2}$ & ... & $\infty$ & $s_{n-1, n}$\\
 n & $s_{n, 0}$ & $s_{n, 1}$ & $s_ {n, 2}$ & ... & $s_{n, n-1}$ & $\infty$
 \end{tabular}
 \label{table-synergy-matrix}
 \end{table}

\subsection{Task planning as multi-objective search}
\label{subsec: synthesis}

The synthesis of a plan $\pi$ uses a domain-independent refinement search, briefly described in Algorithm~\ref{alg: synthesis}.
Timelines are refined iteratively to solve inconsistencies.
We detect flaws in the current partial plan at each iteration, select which flaw to solve, and refine the plan by applying possible solutions.
Each solution determines an alternative refinement and, thus, an alternative partial plan. 
Unexplored partial plans compose the fringe of the search space and are collected into a dedicated data structure in case of backtracking.
A solution is found when the partial plan extracted from the fringe does not contain flaws.

Three points are crucial in Algorithm \ref{alg: synthesis}: first, the selection of the flaw to solve for plan refinement (line 5); second, the selection of the next partial plan (line 8); third, the computation of the objective functions.

\subsubsection{Flaw selection and refinement}

we refer to ``flaws'' as conditions that affect the completeness or validity of timelines.
Flaws may concern tokens to be added to timelines (planning flaws) or tokens of a timeline to be ordered because they overlap (scheduling flaws).
 This choice determines the way the solving process interleaves planning and scheduling decisions.
We implement a hierarchy-based heuristic that considers the synchronization rules of a domain specification 
\cite{epsl-aixia15}\footnote{Intuitively, a higher hierarchical level is assigned to SVs appearing in the rules' trigger (\ie, SVs influencing behaviors of other SVs). A lower hierarchical level is assigned to SVs appearing in the rules' body (\ie, SVs whose behavior depends on other SVs).}.

\subsubsection{Refinement of partial plans}

Given the multi-objective nature of the problem, we pursue a 
Pareto optimality approach \cite{pareto} and apply the {\em dominance relationship} to partial plans.

\begin{definition}
\label{def-dominance}
Given a set $\{f_1, \dots, f_n\}$ of cost functions, a partial plan $\pi_i$ {\em dominates} a partial plan $\pi_j$ (with $i \neq j$) if: 
\[
 f_k(\pi_i) < f_k(\pi_j) \quad\forall\ k=1, ..., n.
\] 
%
\end{definition}

\begin{algorithm}[tpb]
	\caption{Timeline-based plan synthesis}
	\label{alg: synthesis}
	\begin{algorithmic}[1]
		\renewcommand{\algorithmicrequire}{\textbf{Input:}}
		\renewcommand{\algorithmicensure}{\textbf{Output:}}
		\REQUIRE $SV$, $R$, $S^{RH}$
		\ENSURE $\pi =(FTL, R)$
		\STATE $\Pi\ \leftarrow\ \emptyset$
		\STATE $\pi\ \leftarrow\ $ \emph{initialize}$(SV, R)$
		\WHILE {$\neg$ \emph{isSolution}$(\pi, SV, R)$}
		\STATE $\Phi\ \leftarrow\ $\emph{flaws}$(\pi, SV, R)$
		\STATE $\Phi^{*} \leftarrow\ $ \emph{chooseFlaws} $(\Phi, R)$
		\FOR {$\phi\ \in \Phi^{*}$}
		\STATE $\Pi\ \leftarrow\ $ \emph{refine} $(\pi, \phi.\text{resolvers} )$
		\ENDFOR
		\STATE $\pi\ \leftarrow\ $ \emph{choosePlan} $(\Pi, S^{RH})$
		\ENDWHILE
		\RETURN $\pi$
	\end{algorithmic}
\end{algorithm}

\noindent The dominance condition is used to compare partial plans to heterogeneous objective functions and identify the Pareto set of the search space. 
The Pareto set comprises partial plans, representing a suitable trade-off between objective functions $f_k$. 
Such partial plans are compared on a priority assigned to the objectives.
In this work, the objectives are makespan, $f_d$, and synergy, $f_s$. Among the dominant plans, we prioritize synergy.

\subsubsection{Cost estimation of partial plans}
%
The implementation of \eqref{eq-multi-obj} according to the timeline framework needs some intermediate steps.

The objective function must be composed both of a ``cost term'' and a ``heuristic term''. 
Specifically, the cost term models the scheduled tokens of the timelines of a plan $FTL_i \in\pi$. 
Instead, the heuristic term considers possible {\em projections} of the timelines. A projection $\xi^i_j$ represents a particular sequence of tokens $x_k \in \xi^i_j$ that may complete the timeline $FTL_i$ in future refinements of a plan $\pi$. 
All the possible projections of the timeline $FTL_i$ define a set $\Xi_i \ni \xi^i_j$ for all $j$.
Therefore, the minimization passes through the computation of partial plans whose timelines are not necessarily complete.

Consider \eqref{eq-duration-function}, the objective function $f_d\left(\pi\right)$ turns in
\begin{equation}
\label{eq-makespan-plan}
f_d(\pi) = \max_{FTL_i \in\pi} \left( \sum_{x_j \in FTL_i} \!\!d_j
+ \max_{\xi^i_j \in \Xi_i} \sum_{x_j \in \xi^i_j} d_k \right)
\end{equation}
For each timeline $FTL_i$, the makespan turns into the sum of the duration, $d_j$, of its tokens $x_j \in\ FTL_i$ with the maximum sum of the duration $d_k$ of tokens $x_k$ belonging to the projections $\xi^i_j \in \Xi_i$.

Consider \eqref{eq-synergy-function}, the objective function $f_s\left(\pi\right)$ turns in
\begin{equation} \label{eq-synergy-plan}
  f_s(\pi) = \sum_{x^R_i \in FTL_R} \,\sum_{x^H_j \in \Omega(x^R_i)} s_{i, j} 
   + \max_{\xi^R_j \in \Xi_R}\, \sum_{x^R_k \in \xi^R_j} s_{k, *}
\end{equation}
where $FTL_R$ is the robot timeline, $\Omega(x^R_i) = \{x^H_1,\ldots, x^H_n\}$ the set of tokens of the human timeline $FTL_H$ whose execution may overlap in time with $x^R_i$, and the synergy term $s_{i, j}=S^{RH} [x^R_i, x^H_j]$ is extracted from the matrix $S^{RH}$ storing all the synergy terms and it is computed for each pair of overlapping tokens, $x^R_i$ and $x^H_j$. 
Specifically, the first term of \eqref{eq-synergy-plan} is the cost term, while the second one is the heuristic term. This last term considers possible projections of the robot timeline $\xi^R_j \in \Xi_R$ as the maximum expected synergy of the plan according to the worst synergy $S^{RH} [x^R_k, *]$ of the tokens $x^R_k \in \xi^R_j$.

\subsection{Action planning as a multi-goal motion planning problem}
\label{subsec: multi-goal-informed-sampling}
According to Section \ref{subsec: action-planning}, the optimal action planning problem consists of finding the shortest path from the current configuration $q_0 \in U_{g, 0}$ to a vertex in $U_{g, n}$ on a graph $\mathcal{G}=(U_g,E_g)$.
Each edge in $E_g$ corresponds to a motion planning problem between the configurations associated with the edge vertices.

Solving the shortest path on $\mathcal{G}$ is often not viable because evaluating all the edge weights is time-consuming (solving a single motion planning problem may take seconds in a realistic scenario).
To achieve online implementation, we pursue an approximated approach.
We decompose the action planning problem into a sequence of sub-problems, \ie, one sub-problem for each \RobotMovement\ of the \Action.
Then, we optimize the sequence of \RobotMovements\ step by step.
The procedure is described in Algorithm \ref{Algoritmo1}.

Procedure \emph{getGoalsFromScene} gets the \RobotConfigurations, $U_{g,1},\dots,U_{g,n}$ (line 2), where $U_{g, i}$ is the set of configurations goals associated to the $i$th movement.
Then, the algorithm solve a motion planning problem from $q_{\mathrm{current}}$ to $U_{g, i}$ (line 4).
The resulting motion plan is a curve $\sigma: [0, 1] \rightarrow \mathcal{C}$ such that $\sigma(0) = q_{\mathrm{current}}$ and $\sigma(1) \in U_{g, i}$.
The curve $\sigma$ is appended to the array $\mathrm{motion\_plans}$ (line 8), and the total cost is updated (line 9).
Finally, the current configuration is updated with the final configuration of the chosen trajectory (line 10), and the procedure is repeated.

Function \emph{getGoalsFromScene} queries the database containing all locations in the environment. 
If the locations are expressed as Cartesian poses, the function converts them into \RobotConfigurations\ by applying inverse kinematics. Simplifying, the function gets all the configuration goals associated with each movement of an \Action.

\begin{algorithm}[tpb!]
	\caption{Refining tasks into motion plans}
	\label{Algoritmo1}
	\begin{algorithmic}[1]
		\renewcommand{\algorithmicrequire}{\textbf{Input:}}
		\renewcommand{\algorithmicensure}{\textbf{Output:}}
		\REQUIRE action\_model, robot\_tree
		\ENSURE motion\_plans
		\STATE $q_{\mathrm{current}}\leftarrow$ \emph{getCurrentConfiguration}()
		\STATE $(U_{g,1},\dots,U_{g,n})\leftarrow$\emph{getGoalsFromScene}(action\_model, 
		\hspace{160pt} robot\_tree)
		\FOR {$U_{g,i}$ \textbf{in} $(U_{g,1},\dots,U_{g,n})$}
   \STATE $( \sigma, c )$ $\leftarrow$ \emph{motionPlanning}\big(robot\_tree, $q_{\mathrm{current}}$, $U_{g,i}$\big)
   \IF {\emph{isEmpty($\sigma$})} \STATE \textbf{break}
   \ELSE
    \STATE motion\_plans.\emph{append}($\sigma$)
    \STATE cost $\leftarrow$ cost$+c$
    \STATE $q_{\mathrm{current}}\leftarrow\sigma(1)$
   \ENDIF
 	\ENDFOR
 	\RETURN motion\_plans
 	\end{algorithmic}
\end{algorithm}

This approximated approach turns a sub-action into a multi-goal motion planning problem (line 4), for which efficient solvers exist.
Procedure \emph{motionPlanning} outputs the minimum-cost path from a starting configuration to any configuration in $V_{g, i}$.
This problem may be solved by running the motion planner for each configuration in $V_{g, i}$ and by selecting the best solution, although this approach being inefficient and does not scale well to the number of goals.
Informed sampling-based planners \cite{Gammel-InformedRRT} solve the problem efficiently in a single query. 

\section{Qualitative Assessment}
\label{sec: qualitative-assessment}
Existing TAMP approaches for HRC fit different requirements and assumptions, leading to the lack of a shared standard. A quantitative, fair comparison between existing works is difficult because each method is designed to comply with different constraints.
In this section, we resort to examples of real-world problems to argue that our methodology can address a wider variety of cases than existing methods.

\subsection{Case studies}

We consider a typical end-of-line packaging application, where two collaborative lightweight robots are installed, and a human operator can access the cell \cite{Tekniker-pickplace}. 
The robots must pick the packs from a ball-transfer table and place them into boxes. 
Storing packs in the boxes must follow certain rules (\emph{e.g.}, a mosaic composition). 
If needed, the human inspects the packing quality and handles the packs.
The packs on the ball-transfer table are randomly positioned. 
An external camera and an eye-in-hand camera give a raw and refined localization.
In this scenario, we analyze the effectiveness of the most relevant works mentioned in Section \ref{sec: related-works}. 
The results are summarized in Table \ref{tab: comparison}.

\newcommand*\rot{\rotatebox{90}}

\begin{table}
\centering
\caption{Comparison between the proposed TAMP approach and other works in the literature.}
\label{tab: comparison}
\resizebox{.99\columnwidth}{!}{%
\begin{tabular}{r|c|c|c|c|c|c|c|c|c|c|c|}
\multicolumn{1}{c|}{} &
 \multicolumn{1}{c|}{\begin{tabular}[c]{@{}c@{}}\rot{Logic
 constraints}\end{tabular}} &
 \multicolumn{1}{c|}{\begin{tabular}[c]{@{}c@{}}\rot{Geometric constraints}\end{tabular}} &
 \multicolumn{1}{c|}{\begin{tabular}[c]{@{}c@{}}\rot{Temporal constraints}\end{tabular}} &
 \multicolumn{1}{c|}{\begin{tabular}[c]{@{}c@{}}\rot{Hierarchical}\end{tabular}} &
 \multicolumn{1}{c|}{\begin{tabular}[c]{@{}c@{}}\rot{Task optimization}\end{tabular}} &
 \multicolumn{1}{c|}{\begin{tabular}[c]{@{}c@{}}\rot{Optimal robot traj.}\end{tabular}} &
 \multicolumn{1}{c|}{\begin{tabular}[c]{@{}c@{}}\rot{Optimal robot actions}\end{tabular}} &
 \multicolumn{1}{c|}{\begin{tabular}[c]{@{}c@{}}\rot{Contingency strategies} \end{tabular}} &
 \multicolumn{1}{c|}{\begin{tabular}[c]{@{}c@{}}\rot{Robust w.r.t. duration}\end{tabular}} &
 \multicolumn{1}{c|}{\begin{tabular}[c]{@{}c@{}}\rot{Online geom. reasoning}\end{tabular}} \\ \hline
%
\cite{srivastava2014combined}, \cite{Lozano-Perez2014}, \cite{Dantam-incremental-task-motion}, \cite{Lozano-Perez-IJRR2018} & \checkmark\ & \checkmark\ & & \checkmark\ & & & & & & \\ \hline
\cite{Zhang2016}, \cite{Toussaint-logic-geometric} & \checkmark\ & \checkmark\ & & \checkmark\ & \checkmark\ & \checkmark$^1$\ & \checkmark$^1$\ & & & \\ 
\hline
\cite{Magazzeni-temporal-reasoning} & \checkmark\ & \checkmark\ & \checkmark\ & \checkmark\ & & & & & & \\ 
\hline
\cite{Alami-HATP, Alami-Lallement-HATP} & \checkmark\ & \checkmark\ & & \checkmark\ & & & & & & \\ \hline
\cite{Alami-using-human-knowledge, Alami-dealing-with-online-human} & \checkmark\ & \checkmark\ & & \checkmark\ & & & & \checkmark\ & & \\ 
\hline
\cite{Casalino-interleaved-tamp, Casalino-T-RO}& \checkmark\ & \checkmark\ & & \checkmark\ & \checkmark\ & & & \checkmark\ & & \\ 
\hline
\cite{Rosell-knowledge-oriented-tamp, Rosell-Applied-Sciences}& \checkmark\ & \checkmark\ & & \checkmark\ & & & & \checkmark\ & & \\ 
\hline
\cite{Pellegrinelli2017} & \checkmark\ & \checkmark\ & \checkmark\ & \checkmark\ & \checkmark\ & \checkmark$^1$\ & & \checkmark\ & \checkmark\ & \\ 
\hline
Our approach & \checkmark\ & \checkmark\ & \checkmark\ & \checkmark\ & \checkmark\ & \checkmark\ & \checkmark\ & \checkmark\ & \checkmark\ & \checkmark\ \\ 
\hline
\end{tabular}%
}
\begin{flushleft}
\scriptsize $^1$ only offline and if an optimal motion planner is used.
\end{flushleft}
\end{table}

\subsubsection{Temporally bounded process execution}\label{exp: max_time}

Consider the non-collaborative case (\ie, no human intervention) in which the robot cannot overcome a maximum execution time owing to plant constraints.
The methods in Table \ref{tab: comparison} that do not model \textit{temporal constraints} cannot guarantee the plan constraints. This could not be granted even if they integrate \textit{optimal robot trajectories} and \textit{optimal robot actions} because action optimality does not imply the plan is minimum-time. 

\subsubsection{Execution in dynamic environments}\label{exp: dynamic_environmet}

Consider the non-collaborative case in which the position of the remaining packs can change after each grasping operation.
The methods in Table \ref{tab: comparison} that do not implement any \textit{contingency strategies} or \textit{online geometric reasoning} need replanning when an unforeseen event occurs. 
Suppose an action takes longer than expected (\emph{e.g.}, the camera takes longer to identify the grasping position). In that case, \emph{temporal constraints} as in \cite{Magazzeni-temporal-reasoning} may be violated unless the plan is robust to action duration.
Conversely, the algorithms with \textit{contingency strategies} or \textit{online geometric reasoning} need a local replanning.

\subsubsection{Execution with multiple agents}\label{exp: two_agents}
In the case that two robots are used, a delay in the execution of a task (\emph{e.g.}, due to slow processing of a sensor, occlusions of a camera, \textit{etc.}) may introduce synchronization issues between the agents.

All the methods in Table \ref{tab: comparison} that integrate neither \textit{temporal constraints} nor \textit{contingency strategies} can be highly inefficient in the execution.
On the one hand, methods that integrate only \textit{contingency strategies} could cope with the coordination of multiple agents, but a delay in the task execution may violate precedence constraints.
On the other hand, methods that integrate only \textit{temporal constraints} \cite{Magazzeni-temporal-reasoning} may fail in the execution when a worker overcomes the time limits due to unmodeled events (\emph{e.g.}, occlusions, the hold of the movement for safety reasons, \textit{etc.}).
Increasing temporal constraints to compensate for this behavior is not helpful since this information is used only in the planning phase and not during the execution. 
This issue can be mitigated if the temporal modeling used in the plan computation is robust to the task execution latency.

\subsubsection{Optimal execution with multiple agents}
Consider that throughput must be maximized to guarantee the economic return on the robotic cell.
The methodologies in \cite{Toussaint-logic-geometric, Zhang2016} compute an optimal initial plan. 
However, synchronization issues would arise when a task takes longer than expected, as discussed in Section \ref{exp: two_agents}. 
Consequently, a delay in grasping (\eg, due to slow camera perception) leads to an optimality loss or the need to re-compute the whole plan.

\subsubsection{Optimal execution in dynamic environments}
Consider throughput as the goal, and consider that the packs' position can change after each grasping.
\cite{Toussaint-logic-geometric, Zhang2016} cannot grant the plan's feasibility because the trajectory must be recomputed online without excessive idle times.
Although \cite{Pellegrinelli2017} exploits the timelines, it fails because the optimal plan is computed offline, based on a probabilistic model. 
The plan computation should be continuously updated to overcome this limitation in parallel with the task execution. 
However, the latency of the continuous update is high due to the high computational load of the methodology.
\cite{Casalino-interleaved-tamp, Casalino-T-RO} grant local optimality, but movable objects in the scene doe not allow for global optimization.
Finally, \cite{Rosell-knowledge-oriented-tamp, Rosell-Applied-Sciences} embody a contingency strategy to overcome failures related to misalignment between models and reality. 
These methods do not generalize to temporal constraints  and do not allow for trajectory and action optimization.

\subsection{Discussion}

According to the  analysis above, our TAMP approach is the most adequate to deal with typical real-world requirements. 
The capability of considering execution issues at both task and action/motion planning levels is a crucial advantage. This capability enables reliable coordination of agents while preserving the optimality of task assignment, action implementation, and the resulting collaboration.

The combination of temporal flexibility, optimal task allocation, and optimal online motion trajectories allows us to preserve coordination and production efficiency through reliable task plans and behaviors.
Specifically, timeline-based planning effectively integrates reasoning on allocating tasks to the agents and optimizing production while considering possible deviations at execution time. Furthermore, the action planner evaluates the state of the environment online and computes optimal trajectories of motions on the fly. 

\section{Experimental Assessment}
\label{sec: case_study}

\subsection{Case Study}
\begin{figure*}
	\centering
 \begin{minipage}[b]{0.39\textwidth}
 \centering
	\def\lato{0.4}
	\begin{tikzpicture}[every node/.style={minimum size=\lato cm-\pgflinewidth, outer sep=0 pt}]
	\draw[step=\lato cm,color=black] (-2*\lato,1*\lato) grid (0*\lato,3*\lato);
	\node[fill=blue] at (-1.5*\lato,+2.5*\lato) {}; 
	\node[fill=orange] at (-1.5*\lato,+1.5*\lato) {};
	\node[fill=white] at (-0.5*\lato,+2.5*\lato) {}; 
	\node[fill=blue  ] at (-0.5*\lato,+1.5*\lato) {};
	\node at (-2.5*\lato,+1.5*\lato) {2};
	\node at (-2.5*\lato,+2.5*\lato) {1};
	\node at (-1.5*\lato,+3.5*\lato) {A};
	\node at (-0.5*\lato,+3.5*\lato) {B};
\end{tikzpicture}
\,
\begin{tikzpicture}[every node/.style={minimum size=\lato cm-\pgflinewidth, outer sep=0 pt}]
	\draw[step=\lato cm,color=black] (-2*\lato,0*\lato) grid (1*\lato,3*\lato);
	\node[fill=blue] at (-1.5*\lato,+2.5*\lato) {}; 
	\node[fill=orange] at (-1.5*\lato,+1.5*\lato) {};
	\node[fill=orange] at (-1.5*\lato,+0.5*\lato) {};
	\node[fill=blue] at (-0.5*\lato,+2.5*\lato) {}; 
	\node[fill=blue  ] at (-0.5*\lato,+1.5*\lato) {};
	\node[fill=blue] at (-0.5*\lato,+0.5*\lato) {};
	\node[fill=white] at (0.5*\lato,+2.5*\lato) {}; 
	\node[fill=white  ] at (0.5*\lato,+1.5*\lato) {};
	\node[fill=blue] at (0.5*\lato,+0.5*\lato) {};
	\node at (-2.5*\lato,+0.5*\lato) {3};
	\node at (-2.5*\lato,+1.5*\lato) {2};
	\node at (-2.5*\lato,+2.5*\lato) {1};
	\node at (-1.5*\lato,+3.5*\lato) {A};
	\node at (-0.5*\lato,+3.5*\lato) {B};
	\node at (0.5*\lato,+3.5*\lato) {C};
\end{tikzpicture}
\,
\begin{tikzpicture}[every node/.style={minimum size=\lato cm-\pgflinewidth, outer sep=0 pt}]
	\draw[step=\lato cm,color=black] (-2*\lato,-1*\lato) grid (2*\lato,3*\lato);
	\node[fill=orange] at (-1.5*\lato,+2.5*\lato) {}; 
	\node[fill=blue] at (-1.5*\lato,+1.5*\lato) {};
	\node[fill=orange] at (-1.5*\lato,+0.5*\lato) {};
	\node[fill=orange  ] at (-1.5*\lato,-0.5*\lato) {};
	\node[fill=blue] at (-0.5*\lato,+2.5*\lato) {}; 
	\node[fill=blue  ] at (-0.5*\lato,+1.5*\lato) {};
	\node[fill=blue] at (-0.5*\lato,+0.5*\lato) {};
	\node[fill=blue  ] at (-0.5*\lato,-0.5*\lato) {};
	\node[fill=white] at (0.5*\lato,+2.5*\lato) {}; 
	\node[fill=white  ] at (0.5*\lato,+1.5*\lato) {};
	\node[fill=white] at (0.5*\lato,+0.5*\lato) {};
	\node[fill=blue] at (0.5*\lato,-0.5*\lato) {};
	\node[fill=blue  ] at (1.5*\lato,+2.5*\lato) {}; 
	\node[fill=blue  ] at (1.5*\lato,+1.5*\lato) {};
	\node[fill=blue  ] at (1.5*\lato,+0.5*\lato) {};
	\node[fill=white  ] at (1.5*\lato,-0.5*\lato) {};
	\node at (-2.5*\lato,-0.5*\lato) {4};
	\node at (-2.5*\lato,+0.5*\lato) {3};
	\node at (-2.5*\lato,+1.5*\lato) {2};
	\node at (-2.5*\lato,+2.5*\lato) {1};
	\node at (-1.5*\lato,+3.5*\lato) {A};
	\node at (-0.5*\lato,+3.5*\lato) {B};
	\node at (0.5*\lato,+3.5*\lato)  {C};
	\node at (1.5*\lato,+3.5*\lato)  {D};
\end{tikzpicture}
\\ 
\bigskip
\begin{tikzpicture}[every node/.style={minimum size=\lato cm-\pgflinewidth, outer sep=0 pt}]
	\draw[step=\lato cm,color=black] (-2*\lato,-2*\lato) grid (8*\lato,3*\lato);
	\node[fill=orange] at (-1.5*\lato,+2.5*\lato) {}; 
	\node[fill=orange] at (-1.5*\lato,+1.5*\lato) {};
	\node[fill=orange] at (-1.5*\lato,+0.5*\lato) {};
	\node[fill=blue  ] at (-1.5*\lato,-0.5*\lato) {};
	\node[fill=orange] at (-1.5*\lato,-1.5*\lato) {};
	\node[fill=orange] at (-0.5*\lato,+2.5*\lato) {}; 
	\node[fill=blue  ] at (-0.5*\lato,+1.5*\lato) {};
	\node[fill=orange] at (-0.5*\lato,+0.5*\lato) {};
	\node[fill=blue  ] at (-0.5*\lato,-0.5*\lato) {};
	\node[fill=orange] at (-0.5*\lato,-1.5*\lato) {};
	\node[fill=orange] at (0.5*\lato,+2.5*\lato) {}; 
	\node[fill=blue  ] at (0.5*\lato,+1.5*\lato) {};
	\node[fill=orange] at (0.5*\lato,+0.5*\lato) {};
	\node[fill=orange] at (0.5*\lato,-0.5*\lato) {};
	\node[fill=orange] at (0.5*\lato,-1.5*\lato) {};
	\node[fill=blue  ] at (1.5*\lato,+2.5*\lato) {}; 
	\node[fill=blue  ] at (1.5*\lato,+1.5*\lato) {};
	\node[fill=blue  ] at (1.5*\lato,+0.5*\lato) {};
	\node[fill=blue  ] at (1.5*\lato,-0.5*\lato) {};
	\node[fill=blue  ] at (1.5*\lato,-1.5*\lato) {};
	\node[fill=white ] at (2.5*\lato,+2.5*\lato) {}; 
	\node[fill=white ] at (2.5*\lato,+1.5*\lato) {};
	\node[fill=white ] at (2.5*\lato,+0.5*\lato) {};
	\node[fill=blue  ] at (2.5*\lato,-0.5*\lato) {};
	\node[fill=blue  ] at (2.5*\lato,-1.5*\lato) {};
	\node[fill=blue  ] at (3.5*\lato,+2.5*\lato) {}; 
	\node[fill=blue  ] at (3.5*\lato,+1.5*\lato) {};
	\node[fill=blue  ] at (3.5*\lato,+0.5*\lato) {};
	\node[fill=white ] at (3.5*\lato,-0.5*\lato) {};
	\node[fill=white ] at (3.5*\lato,-1.5*\lato) {};
	\node[fill=blue  ] at (4.5*\lato,+2.5*\lato) {}; 
	\node[fill=white ] at (4.5*\lato,+1.5*\lato) {};
	\node[fill=white ] at (4.5*\lato,+0.5*\lato) {};
	\node[fill=blue  ] at (4.5*\lato,-0.5*\lato) {};
	\node[fill=blue  ] at (4.5*\lato,-1.5*\lato) {};
	\node[fill=blue  ] at (5.5*\lato,+2.5*\lato) {}; 
	\node[fill=blue  ] at (5.5*\lato,+1.5*\lato) {};
	\node[fill=blue  ] at (5.5*\lato,+0.5*\lato) {};
	\node[fill=white ] at (5.5*\lato,-0.5*\lato) {};
	\node[fill=white ] at (5.5*\lato,-1.5*\lato) {};
	\node[fill=white ] at (6.5*\lato,+2.5*\lato) {}; 
	\node[fill=white ] at (6.5*\lato,+1.5*\lato) {};
	\node[fill=white ] at (6.5*\lato,+0.5*\lato) {};
	\node[fill=blue  ] at (6.5*\lato,-0.5*\lato) {};
	\node[fill=blue  ] at (6.5*\lato,-1.5*\lato) {};
	\node[fill=blue  ] at (7.5*\lato,+2.5*\lato) {}; 
	\node[fill=blue  ] at (7.5*\lato,+1.5*\lato) {};
	\node[fill=blue  ] at (7.5*\lato,+0.5*\lato) {};
	\node[fill=blue  ] at (7.5*\lato,-0.5*\lato) {};
	\node[fill=blue  ] at (7.5*\lato,-1.5*\lato) {};
	\node at (-2.5*\lato,-1.5*\lato) {5};
	\node at (-2.5*\lato,-0.5*\lato) {4};
	\node at (-2.5*\lato,+0.5*\lato) {3};
	\node at (-2.5*\lato,+1.5*\lato) {2};
	\node at (-2.5*\lato,+2.5*\lato) {1};
	\node at (-1.5*\lato,+3.5*\lato) {A};
	\node at (-0.5*\lato,+3.5*\lato) {B};
	\node at (0.5*\lato,+3.5*\lato) {C};
	\node at (1.5*\lato,+3.5*\lato) {D};
	\node at (2.5*\lato,+3.5*\lato) {E};
	\node at (3.5*\lato,+3.5*\lato) {F};
	\node at (4.5*\lato,+3.5*\lato) {G};
	\node at (5.5*\lato,+3.5*\lato) {H};
	\node at (6.5*\lato,+3.5*\lato) {I};
	\node at (7.5*\lato,+3.5*\lato) {J};
\end{tikzpicture}
\end{minipage}\hfill
\begin{minipage}[b]{0.6\textwidth}
	\includegraphics[trim = 6cm 18cm 6cm 4.5cm, clip, angle=0, width=0.85\columnwidth]{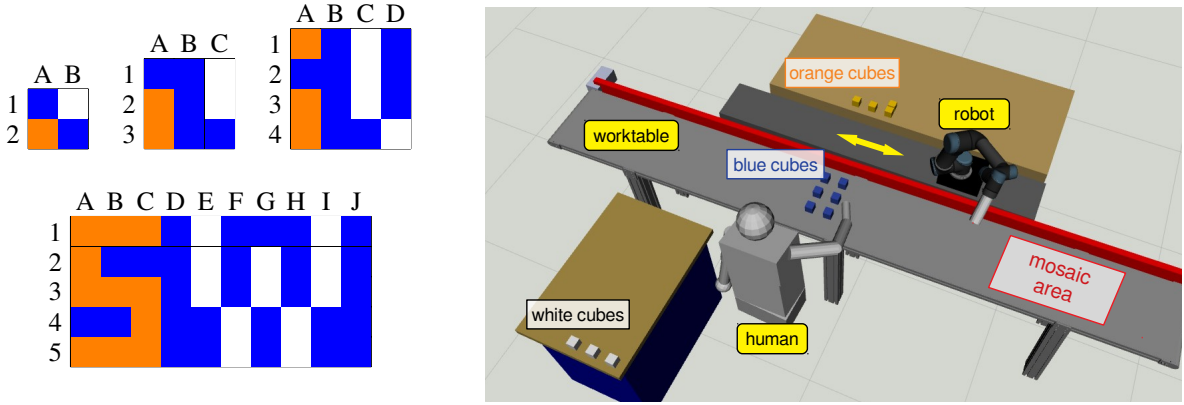}
 \end{minipage}
	\caption{Case study: a 7-degree-of-freedom robot and a human worker collaborate to assemble a mosaic.
	Four mosaics are considered, with the number of cubes ranging from 4 to 50.}
	\label{fig: mosaic}
\end{figure*}

We consider a case study derived from the EU-funded project \emph{ShareWork} (http://www.sharework-project.eu) where a robot arm (Universal Robots UR5 on an actuated linear track) and a human operator have to assemble a mosaic (see Figure \ref{fig: mosaic}). 
We consider four mosaics composed of 4, 9, 16, and 50 cubes of different colors (blue, orange, and white). Each slot has a label given by its column letter and its row number (\ie, \emph{A1, A2, $\ldots$}).
A common condition in HRC is that some operations can be performed only by the robot or the human. 
In this example, the following allocation constraints are imposed:
orange cubes shall be moved only by the robot;
white cubes shall be moved only by the human;
both can move blue cubes.

\subsubsection{Process planning model}
\label{subsec: case-study-task-planner}
The case study is traced back to the timeline-based formalism described in Section \ref{subsec: problem-ps}.
The process is modeled as a production state variable $SV^P = \left(V^P, T^P, D^P, \gamma^P \right)$, where each value of $V^P$ represents the assembly of a row, \ie, $V^P=\{\mathit{DoRow}_1, \mathit{DoRow}_2, \dots \}$.
The precedence of some rows over others may be set at this level as synchronization rules.
Two behavior state variables, $SV^H = \left(V^H, T^H, D^H, \gamma^H \right)$ and $SV^R = \left(V^R, T^R, D^R, \gamma^R \right)$, model the low-level tasks that the human and the robot can perform.
In this case study, the human and the robot perform tasks of the type $\mathit{PickPlace}_x$, which consists of picking a cube and placing it in the slot with label $x$.
Hence, $V^H = \{\mathit{PickPlace}_y\}$ and $V^R = \{\mathit{PickPlace}_z\}$, where $y$ is the labels corresponding to white and blue cubes, and $z$ are labels corresponding to orange and blue cubes. Note that $V^R \cap\ V^H \neq \emptyset$ because blue tiles can be assigned to humans and robots.

According to the controllability notion given in Section \ref{subsec: problem-ps}, the human behavior is modeled as uncontrollable ($\gamma^H(v_k)=u\, \, \forall\, v_k \in\ V^H$) and the robot behavior of is partially controllable ($\gamma^R(v_k)=pc \, \, \forall\, v_k \in V^R$).
The duration of each task ($D^R$ and $D^H$) is estimated as in \cite{pellegrinelli2018estimation}. This duration estimation method considers the human interference on the robot paths and estimates the duration for all possible obstruction and safety stop cases. Based on this estimation, each element $s_{i, j}$ of the synergy matrix is computed as in \eqref{eq-synergy-function}.

Each task $\mathit{PickPlace}_z \in V^R$ corresponds to a set of actions to be performed by the robot.
Each action boils down to a sequence of \RobotMovements, as shown in Figure \ref{fig: action_model}.
When the action planner receives a task, the action planner decodes the cube's color and the goal slot from a database.
Then, it solves the action planning problem described in Algorithm \ref{Algoritmo1}.

\subsubsection{Software implementation}
\label{subsec: sw-implementation}
The task planner in Section \ref{sec: solvers} has been integrated into the timeline-based planner PLATINUm \cite{aixia17}, using a \emph{hierarchy-based heuristics} for flaw selection and an \emph{HR-balancing search strategy} for search expansion \cite{Faroni_ROMAN2020}. The rest of the framework is implemented in ROS and connected to PLATINUm by \emph{rosbridge\_suite} \cite{rosbridge}.
PLATINUm dispatches tasks to the robot action planner and waits for feedback.
In real-world tests, an HMI system would communicate the task to humans and receive feedback from them.
In simulated tests, the human is modeled through a mannequin\footnote{The modeled movements are: trunk (2 translations and 3 rotations), shoulders (3 rotations each), elbow (1 rotation each), and wrist (3 rotations).} commanded by a second instance of the action planner used for the robot\footnote{Using a simulation model of the environment and the human worker is fundamental to ensure high repeatability of the tests, erasing the effects of measurement uncertainty, differences between human subjects, and human-machine communication.
This allows for a fair comparison between different methods, focusing only on the effects of task and motion planning.}.

The action planner is implemented by using the library \cite{Villagrossi-manipulation}, which builds high-level skills on top of \emph{MoveIt!}.
Because of the multi-goal nature of the proposed action planning and the online requirement, we use MI-RRT$^*$ \cite{DIRRT}, a fast variant of Informed-RRT$^*$ \cite{Gammel-InformedRRT}, to solve the motion planning problems.

\subsection{Experiments}
We discuss three experiments to evaluate the different components of the proposed approach. The first two analyze the reasoning capabilities of the task planner and action planner alone. The third one then evaluates their integration within the proposed TAMP approach. We conclude the section with a final discussion of the results emphasizing the main advantages and strengths of the approach.
A video of the experiments is attached to this manuscript.

\subsubsection{Experiment 1 (task planning performance)}
We compare our approach with a timeline-based approach \cite{Pellegrinelli2017} and an action-based approach \cite{Magazzeni-temporal-reasoning} (according to Table \ref{tab: comparison}, \cite{Pellegrinelli2017} and \cite{Magazzeni-temporal-reasoning} are the only approaches that can manage temporal constraints).

\paragraph{Comparison with timeline-based approaches}
\cite{Pellegrinelli2017} has two main limitations.
First, the task planner reasons at a low-level of abstraction (\ie, each point-to-point movement is modeled as a \Task), putting the task planner in charge of finding the optimal ordering and assignment of all the \RobotMovements.
Second, it is based on the a-priori modeling of the trajectories, \ie, all trajectories are pre-computed, and the task planner reasons on the estimated costs of such plans. 

To demonstrate that our approach scales better than \cite{Pellegrinelli2017} to complex processes, we consider the mosaics in Figure~\ref{fig: mosaic}.
\cite{Pellegrinelli2017} models each \RobotMovement\ from a cube to each slot (and vice versa) as a \Task.
On the contrary, the proposed method only requires one \Task\ for each slot.
As the planning time roughly grows exponentially in the number of \Tasks, the planning time of the proposed method is around one order of magnitude smaller than that of \cite{Pellegrinelli2017}, as shown in Figure \ref{fig: comparison-pellegrinelli} (right plot).
Note that \cite{Pellegrinelli2017} could not solve the 50-cube mosaic within the maximum planning time of 15 minutes.

Similar reasoning holds for the motion planning phase. \cite{Pellegrinelli2017} pre-computes all trajectories from all cubes to all slots and vice versa. 
Suppose the number of cubes in the scene is equal to the number of slots of the mosaic, and let $\tau_{\mathrm{max}}$ be the maximum planning time after which a motion planning query is stopped and $b$ the number of slots. Then, the offline phase can take up to $2\tau_{\mathrm{max}}b^2$ seconds.
Considering that, in our tests, $\tau_{\mathrm{max}}=5$~s, this corresponds to 90, 360, 1690, and 14440 seconds for each mosaic.
On the contrary, our approach computes the trajectories online, dealing with uncertainty and changing goals.

Concerning the execution phase, we compare the performance of the two approaches for the 4-cube, 9-cube, and 16-cube mosaics.
Figure \ref{fig: comparison-pellegrinelli} shows that the execution time of the proposed method is significantly shorter (-14\%, -12\%, -20\% for the three mosaics) because the action planner chooses the most convenient movement online and based on the current robot and human state. Indeed, the robot's traveled distance is much shorter (-26\%, -39\%, -27\% for the three mosaics).
This difference is even more evident when more cubes than necessary are available.
For example, we consider the case where 50 cubes are available although only 4, 9, and 16 are necessary (method \vv{proposed w/ all cubes} in Figure \ref{fig: comparison-pellegrinelli}). 
Because our method can choose among a broad set of objects at each $\mathit{PickPlace}_x$ task, it leads to a further improvement of the execution time (-44\%, -22\%, -34\% for the three mosaics) and the robot's traveled distance (-63\%, -39\%, -46\% for the three mosaics).

\paragraph{Comparison with action-based approaches}
A direct comparison with other approaches mentioned in Table \ref{tab: comparison} is difficult because of the intrinsic differences in terms of planning formalism and models. 
%
Nonetheless, we consider \cite{Magazzeni-temporal-reasoning} as it supports {\em temporal constraints} and {\em hierarchical decomposition}. 
The comparison thus focuses on how the two different models deal with uncertain and strict temporal requirements.

\cite{Magazzeni-temporal-reasoning} proposes an integration of task and motion planning capabilities based on PDDL2.1 \cite{pddl21}. 
It uses an action-based representation with so-called {\em durative actions} that do not consider scheduling aspects. 
PDDL2.1 planners do not pursue makespan optimization but use temporal constraints for plan consistency.
Furthermore, neither {\em temporal flexibility} nor {\em temporal uncertainty} are considered by such planners.
Therefore, the task planning models pick-and-place tasks with a fixed duration and consider the robot and the worker as controllable.
We run our implementation of \cite{Magazzeni-temporal-reasoning} for the 4-cube, 9-cube, and 16-cube mosaic and show the results in Table \ref{tab: comparison_magazzeni}.
As expected, \cite{Magazzeni-temporal-reasoning} achieves the best planning time for all scenarios because the action-based planner focuses on process decomposition without considering optimization aspects. 
Scheduling decisions do not impact the reasoning, while the lack of flexibility reduces the number of choices considered during the search.
Then, we execute the plans obtained for the 4-cube, 9-cube, and 16-cube mosaic simulating the uncertainty of the worker's actions ($\delta\ = \pm\ 5$ time units). 
The objective is to evaluate the reliability of synthesized plans in a realistic scenario where human workers behave uncontrollably.
As shown in Table \ref{tab: comparison_magazzeni}, the execution time of \cite{Magazzeni-temporal-reasoning} is greater than the execution of our approach (up to +110\%). 
The lack of temporal flexibility with respect to uncertainty does not allow to deal with the uncontrollable dynamics of the worker effectively. Consequently, the frequent need for re-planning increases the execution time of the plan, leading to less efficient (and effective) collaborations.

\begin{table}[tpb]
\caption{Results of Experiment 1.2. Comparison of the proposed method and \cite{Magazzeni-temporal-reasoning}. }	
\label{tab: comparison_magazzeni}
\centering
$
\begin{array}{lcccc}
\toprule
&     & \textbf{Proposed} & \text{Edelkamp et al. \cite{Magazzeni-temporal-reasoning}} \\
 \midrule
 \multirow{4}{*}{\text{Planning time [ms]}} & \text{4 cubes} & \textbf{392}(6.1)  & 263(34) \\
            & \text{9 cubes} & \textbf{1991}(240) & 947(91)  \\
            & \text{16 cubes} & \textbf{8942}(176) & 5099(756) \\
 \midrule
	\multirow{4}{*}{\text{Execution time [s]}} & \text{4 cubes} & \textbf{43}(5.7) & 65(6.1) \\
            & \text{9 cubes} & \textbf{85}(6.3) & 180(8.5) \\
            & \text{16 cubes} & \textbf{166}(7.2) & 197(6.1) \\
 \bottomrule
	\end{array}
$
\end{table}

\subsubsection{Experiment 2 (action planning performance)}
\label{subsec: experiment2}
Consider the 50-cube mosaic of Figure~\ref{fig: mosaic} and the following action-planning configurations:
\begin{enumerate}[label=\roman*.]
	\item \textbf{pre-computed}: motion plans from and to each point are computed offline, as in \cite{Pellegrinelli2017}. Before each movement, the algorithm chooses the closest goal to the current robot position.
	Referring to Table \ref{tab: comparison}, this approach displays \emph{optimal robot trajectories}, but neither \emph{optimal actions} or \emph{online geometric reasoning}.
	It cannot deal with dynamic environments as paths are computed \emph{a priori}.
	\item \textbf{single-goal}: motion plans are computed before execution. The action planner always selects the closest goal to the current robot position.
	Referring to Table \ref{tab: comparison}, this approach has \emph{optimal robot trajectories} and \emph{online geometric reasoning}, but no \emph{optimal actions}.
	\item \textbf{multi-goal}: the proposed action planning. Motion plans are computed just before their execution. The multi-goal optimal motion planner considers all the goals with the desired properties, yielding \emph{optimal robot trajectories}, \emph{online geometric reasoning}, and \emph{optimal actions}.
\end{enumerate}

We evaluate the following indexes: 
i) total execution time of the robot tasks (in seconds); 
ii) joint-space distance traveled by the robot (in radians); 
iii) planning time for the motion planning algorithm, \ie, the sum of the planning times of all movements to perform a pick-and-place action (in seconds).

We run 20 tests for each configuration by using a task plan generated by the \textbf{feasible} configuration described in Section \ref{subsec: experiment1} (the chosen plan assigns 25 tasks to the robot and 25 tasks to the human).
Results are in Figure~\ref{fig: motion-results}. 
The \textbf{multi-goal} configuration outperforms the \textbf{pre-computed} and the \textbf{single-goal} variants with a reduction of around 48\% in the traveled distance (\textbf{pre-computed}: mean = 1487 rad, stdev = 23.7 rad; \textbf{single-goal}: mean = 1480 rad, stdev = 27.1 rad; \textbf{multi-goal}: mean = 768 rad, stdev = 2.21 rad).
All configurations use the same optimal path planner; the improvement of the solution is due to the choice of the goal: \textbf{pre-computed} and \textbf{single-goal} direct the search towards the closest goal, which is often sub-optimal. 
Other heuristics could be adopted to select the goal, but the results would strongly depend on the geometric properties of the workspace. 
For example, the \vv{closest-one} heuristic would perform even worse in a cluttered environment.
On the contrary, \textbf{multi-goal} always finds the best solution regardless of the geometry of the workspace.
Shorter paths reduce the execution time, as shown in Figure~\ref{fig: motion-results}.
Note that the difference in the execution time is less pronounced than the one in the traveled distance. 
The reason is that the execution time also considers planning latency, safety slowdown, and communication overhead, and it is affected by the path parametrization algorithm.
Nonetheless, \textbf{multi-goal} leads to a significant reduction (around 21\%) in the robot execution time (\textbf{pre-computed}: mean = 632 s, stdev = 12.0 s; \textbf{single-goal}: mean = 629 s, stdev = 8.05 s; \textbf{multi-goal}: mean = 500 s, stdev = 1.27 s).
Note that the \textbf{multi-goal} approach results in a minor variance of the traveled distance and execution time because the multi-goal search is less affected by the robot state at the beginning of each action. 

Finally, the motion planning times are shown in the right plot of Figure~\ref{fig: motion-results} (\textbf{pre-computed} is not shown as all paths are computed offline). 
As expected, \textbf{multi-goal} has planning times higher than \textbf{single-goal} (\textbf{single-goal}: mean = 0.567 s, stdev = 0.057 ms; \textbf{multi-goal}: mean = 2.65 s, stdev = 0.211 s).
This difference is intrinsic to the multi-goal nature of the planner, which has to plan toward all available goals.
Nonetheless, planning times of \textbf{multi-goal} are still in the order of a few seconds and, therefore, suitable for online planning. 
Moreover, this discrepancy is expected to decrease thanks to the constant advances in multi-goal motion planning algorithms.

\subsubsection{Experiment 3 (TAMP performance)}
\label{subsec: experiment1}
Consider the 50-cube mosaic of Figure~\ref{fig: mosaic} and the following cases:
\begin{enumerate}[label=\roman*.]
	\item \textbf{feasible}: we generate feasible random task plans with respect to allocation and precedence constraints but do not optimize duration and synergy. 
	The number of tasks assigned to the human is chosen as a uniform random variable between 12 and 39 (\ie, the smallest and the largest number of cubes that the human can move).
	Referring to Table \ref{tab: comparison}, this configuration reproduces those methods that do not implement any \emph{task optimization}, \emph{contingency strategy}, or \emph{temporal robustness}.
	\item \textbf{optimized}: we use the proposed multi-objective optimization approach. The task planner decides the order and allocation of the tasks.
\end{enumerate}
Both configurations use the proposed action planner to highlight the differences owing to the task plan generation.

For \textbf{optimized}, the task planner decides the number of tasks for each worker in such a way as to minimize the process duration.
This is a key feature of our approach, while existing works either assume that the number of assignments is given or they find a feasible assignment, disregarding its optimality.
To reproduce this issue, we let the \textbf{feasible} approach randomly decide the number of tasks assigned to each worker as long as the assignment is feasible.
The following indexes are evaluated:
\begin{enumerate}[label=\roman*.]
	\item process Execution Time, 
	$\mathrm{ET}_P = \max(\mathrm{ET}_R, \mathrm{ET}_H)$
	\item Idle Time $\mathrm{IT} = 100\, | \mathrm{ET}_R - \mathrm{ET}_H | / \mathrm{ET}_P$ [\%]
	\item Concurrent working Time of human and robot, 
	$\mathrm{CT} = 100\cdot( \min(\mathrm{ET}_R, \mathrm{ET}_H) - \mathrm{ST})/\mathrm{ET}_P$ [\%]
\end{enumerate}
where $\mathrm{ET}_R$ and $\mathrm{ET}_H$ are the execution time of the robot and the human, and $\mathrm{ST}$ is the robot holding time because of safety (\ie, when the human is close to the robot).
The first index measures the throughput of the process, and the second and the third measure the quality of collaboration.

\begin{figure*}[t]
 \centering
 \includegraphics[trim = 3.5cm 0.2cm 3.5cm 0.2cm, clip, angle=0, width=0.9\textwidth]{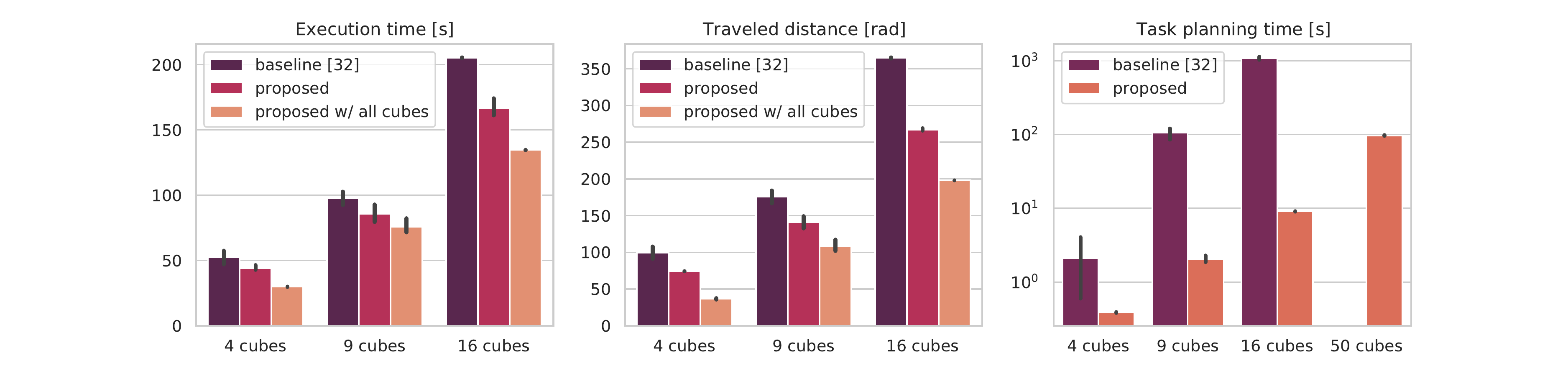}
 \caption{Results of Experiment 1.a. The proposed method (\textbf{proposed}) is compared with \cite{Pellegrinelli2017} (\textbf{baseline}). The \textbf{proposed} method reduces the execution time (left plot), the robot's traveled distance (middle plot), and the planning time of the task planner (right plot). The difference is even more evident when more cubes than strictly necessary are used (\textbf{proposed with all cubes}).}
 \label{fig: comparison-pellegrinelli}
 \centering
 \includegraphics[trim = 3.5cm 0.2cm 3.5cm 0.2cm, clip, angle=0, width=0.9\textwidth]{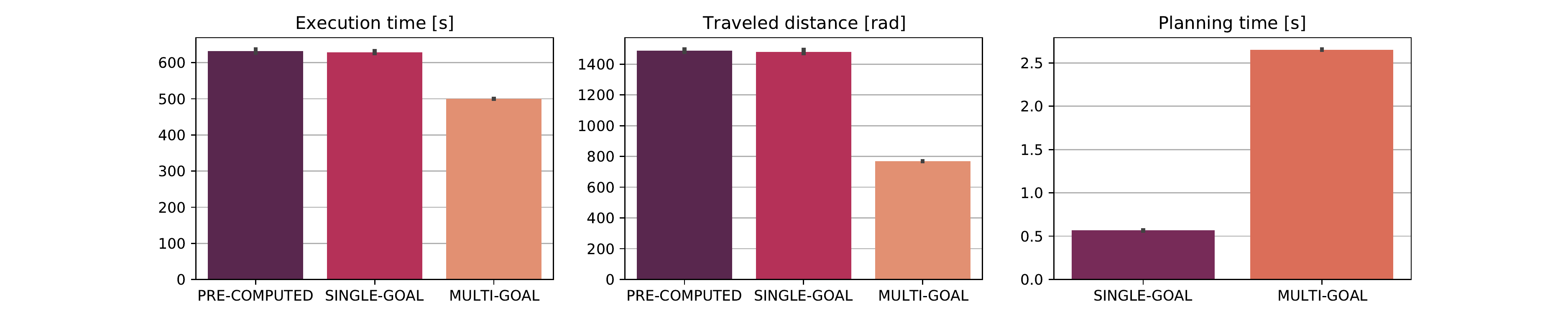}
 \caption{Results of Experiment 2. The proposed \textbf{multi-goal} action planner is compared with a \textbf{pre-computed} motion planner and a \textbf{single-goal} planner.
 The path length is reduced (left plot) as the robot execution time (middle plot). 
 Planning times are larger but suitable for online planning (right plot; \textbf{pre-computed} is not shown because it relies on offline planning).}
 \label{fig: motion-results}
 \centering
 \includegraphics[trim = 3.5cm 0.2cm 3.5cm 0.2cm, clip, angle=0, width=0.9\textwidth]{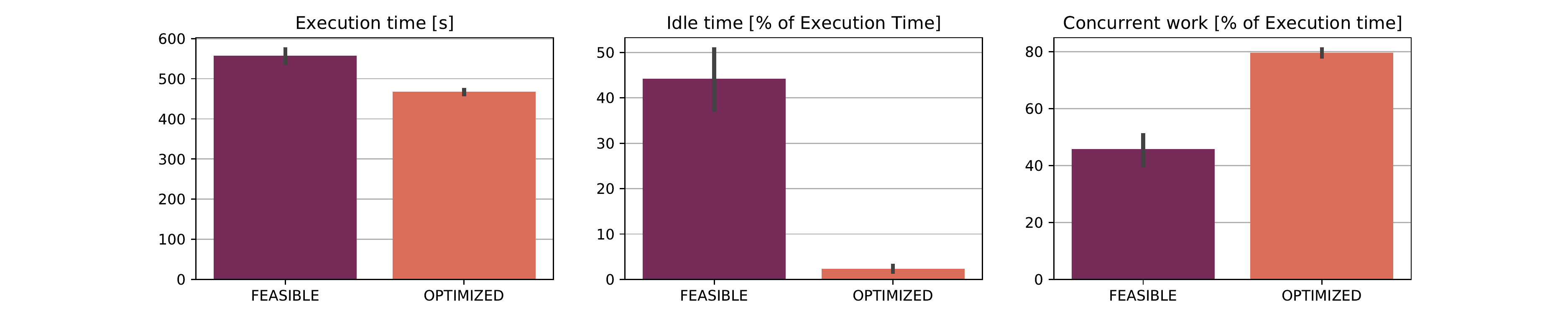}
 \caption{Results of Experiment 3. The proposed method (\textbf{optimized}) is compared with a feasibility-oriented approach (\textbf{feasible}): \textbf{optimized} reduces the process execution time (left plot) and the idle time of human and robot (middle plot) and increases the time the robot and the human work simultaneously (right plot).}
 \label{fig: tamp-results}
\end{figure*}

We run 20 tests for each configuration; results are in Figure~\ref{fig: tamp-results}.
The \textbf{optimized} approach outperforms the \textbf{feasible} one by reducing $\mathrm{ET}_P$ of around 16\% (\textbf{optimized}: mean = 467 s, stdev = 11.9 s; \textbf{feasible}: mean = 557 s, stdev = 39.3 s) and $\mathrm{IT}$ of around 95\% (\textbf{optimized}: mean = 2.36\%, stdev = 1.50\%; \textbf{feasible}: mean = 44.2\%, stdev = 15.2\%), while increasing CT of around 74\% (\textbf{optimized}: mean = 79.6\%, stdev = 3.26\%; \textbf{feasible}: mean = 45.8\%, stdev = 12.1\%).
Note that the \textbf{optimized} approach displays a balanced assignment of tasks to the robot and the human.
In this case, the task planner assigns 27 tasks to the human and 23 tasks to the robot, so the two expected makespans are similar.
As a result, the execution time and the idle time are shorter.

\subsection{Discussion}
The outcomes of these experiments confirm the conclusions of the qualitative assessment. 
The task planning assessment shows that the timeline-based approach achieves a higher level of reliability during plan execution in the case of uncontrollable delays or temporal deviations in the execution of tasks.
It also shows that the proposed approach scales to more complex tasks than previous timeline-based methods (Figure \ref{fig: comparison-pellegrinelli}).

The action planning assessment shows that the dynamic selection of the motion goal significantly increases the flexibility and reliability of robot motions and achieves shorter execution time and robot traveled distance (Figure \ref{fig: motion-results}).

The third experiment focuses on integration. 
It shows the advantages of integrating the two planning approaches. 
The results of Figure \ref{fig: tamp-results} clearly show that the combination of optimal reasoning at the two levels of abstraction significantly improves the synergetic behaviors of the agents in terms of both idle time and concurrency.

\section{Conclusions and future works}
This paper proposed a task and motion planning approach for hybrid collaborative processes.
The proposed method follows a multi-objective optimization approach to maximize the throughput of the process.
%
We demonstrated the advantages of the method compared with state-of-the-art techniques, both from a qualitative and numerical point of view.
Future works will focus on integrating learning techniques to refine the process model through experience and speed up the search for optimal plans \cite{T-CYB-task-learning}.
For example, \cite{Sandrini_ETFA2022} presents a preliminary study on learning task duration and human-robot synergy via linear regression.
Further investigation will also address the use of other optimization objectives (\emph{e.g.}, taking into account human factors and user preferences \cite{Lippi:task-allocation} or agent-agent communication \cite{T-CYB:gaze}). 
Finally, real-world tests of human workers will assess the system's performance and dependability.

\bibliographystyle{IEEEtran}
\bibliography{references, reference_stiima, new_bib}

\end{document}